# Vector Quantization using the Improved Differential Evolution Algorithm for Image Compression

[1]Sayan Nag[*]
[1]Department of Electrical Engineering,
Jadavpur University
Kolkata
India

[*]Corresponding Author

*Abstract*- **Vector Quantization (VQ) is a popular image compression technique with a simple decoding architecture and high compression ratio. Codebook designing is the most essential part in Vector Quantization. Linde–Buzo–Gray (LBG) is a traditional method of generation of VQ Codebook which results in lower PSNR value. A Codebook affects the quality of image compression, so the choice of an appropriate codebook is a must. Several optimization techniques have been proposed for global codebook generation to enhance the quality of image compression. In this paper, a novel algorithm called IDE-LBG is proposed which uses Improved Differential Evolution Algorithm coupled with LBG for generating optimum VQ Codebooks. The proposed IDE works better than the traditional DE with modifications in the scaling factor and the boundary control mechanism. The IDE generates better solutions by efficient exploration and exploitation of the search space. Then the best optimal solution obtained by the IDE is provided as the initial Codebook for the LBG. This approach produces an efficient Codebook with less computational time and the consequences include excellent PSNR values and superior quality reconstructed images. It is observed that the proposed IDE-LBG find better VQ Codebooks as compared to IPSO-LBG, BA-LBG and FA-LBG.**

*Keywords – Image Compression, Vector Quantization, Codebook, Improved Differential Evolution (IDE) Algorithm, Linde-Buzo-Gray (LBG) Algorithm, Improved Particle Swarm Optimization (IPSO) Algorithm, Bat Algorithm (BA), Firefly Algorithm (FA).*

______________________________________

[*]Corresponding Author, Email: nagsayan112358@gmail.com

# 1. Introduction

The major role of Image compression lies in medical sciences, internet browsing and navigation applications, TV broadcasting and so on. With the advancement of science and technology, the storage space requirement and the need for reduction in transmission time for digital images are becoming essential concerns. The transmission bandwidth being limited creates a problem. An efficient and proper image compression technique is of prime requirement to tackle the problem of limited bandwidth. This area of study attracted many researchers over the past decades who have suggested several techniques for image compression. Vector Quantization (VQ) technique outperforms other techniques such as pulse code modulation (PCM), differential PCM (DPCM), Adaptive DPCM which belongs to the class of scalar quantization methods. Vector quantization (VQ) [1, 2], one of the most popular lossy image compression methods is primarily a c-means clustering approach widely used for image compression as well as pattern recognition [3, 4], speech recognition [5], face detection [6] speech and image coding because of its advantages which include its simple decoding architecture and the high compression rate it provides with low distortion- hence its popularity. Vector Quantization involves the following three steps: (i) Firstly, the image in consideration is divided into non-overlapping blocks commonly known as input vectors. (ii) Next, a set of representative image blocks from the input vectors is selected referred to as a codebook and each representative image vector is referred to as a code-word. (iii) Finally, each of the input vectors is approximately converted to a code-word in the codebook, the corresponding index of which is then transmitted.

Codebook training is considered sacrosanct in the process of Vector Quantization, because a codebook largely affects the quality of image compression. Such significance of Codebook training gave new impetus to many researchers leading to the proliferation of researches to design codebook using several projected approaches. Vector Quantization methods are categorized into two classes: crisp and fuzzy [7]. Sensitive to codebook initialization, Crisp VQ follows a hard decision making process. Of this type, C-Means algorithm is the most representative of all. Linde et al. propounded the famous Linde-Buzo-Gray (LBG) algorithm in this respect. Starting with the smallest possible codebook size the gradual parlay in the size occurs by using a splitting procedure [8]. By implanting specific functions (utility measures) in the learning process the performance of the LBG algorithm is significantly enhanced. Fuzzy VQ incorporates fuzzy C-Means algorithm. The approach is similar to that of fuzzy cluster analysis. Each training vector is posited to be a property of multiple clusters with membership values which makes the learning a soft decision making process [9]. The drawback of LBG Algorithm is that it gets stuck in a local optimum. In order to overcome such problem, a clustering algorithm known as enhanced LBG is proposed [10]. By changing the lower utility code-words to the one with higher utility the problem of local optimal with LBG is overcome.

Recently researches involve the association of the evolutionary optimization algorithms with the LGB algorithm. This conglomeration for design the codebook has a significant impact in improving the results of the LGB algorithm. Rajpoot, et al. formulated a codebook using an Ant Colony Optimization (ACO) algorithm [11] yielding fascinating results thereby clearly depicting the improvement in the results obtained by the hybridization of ACO with LBG. Codebook is optimized abreast appointing the vector coefficients in a bidirectional graph; next, a appropriate implementation of placing pheromones on the edges of the graph is expounded in the ACO-LBG Algorithm. The speed of convergence of the ACO_LBG algorithm is proliferated by Tsaia et al. by discarding the redundant calculations [12]. Particle Swarm Optimization (PSO) is also an adaptive swarm optimization approach based on updating the global best (gbest) and local best (lbest) solutions. The ease of improvement and fast convergence to an expected solution attracted many researchers to apply PSO for solving optimization problems. Particle swarm optimization along with vector quantization [13] overcomes the drawbacks of the LBG algorithm. Evolutionary fuzzy particle swarm optimization algorithm [14] is an efficient and robust algorithm in terms of performances compared to that of the LBG learning algorithms. Quantum particle swarm algorithm (QPSO) is yet another PSO put forward by Wang et al. for solving the 0–1 knapsack problem [15]. Applied together with LBG, the QPSO-LBG algorithm outperforms LBG algorithm as well.

Yang et al. devised a new method for image compression which involves an Evolutionary clustering based vector quantization primarily focused on One-Step Gradient Descent Genetic Algorithm (OSGD-GA). The particular algorithm is formulated for optimizing the codebooks of the low-frequency wavelet coefficient by expounding the consequence of every coefficient involved abreast employing fuzzy membership values for automatic clustering [16].

The Firefly Algorithm (FA) is an efficient Swarm Intelligence tool which is largely applied to many engineering design problems nowadays. Firefly Algorithm is inspired by the social activities of fireflies. In presence of a firefly with brighter intensity, the one with lower intensity value move towards the former and in the absence of the brighter firefly then one with lower intensity moves in a random fashion. Horng successfully applied a newly developed Firefly Algorithm (FA) for designing the codebook for vector quantization [17]. An efficient algorithm by the name of Honey Bee Mating Optimization (HBMO) is applied for codebook generation yielding a good quality reconstructed image with modicum of distortion in Image Compression Problems [18]. Dynamic programming is used as an efficient optimization technique for layout optimization of interconnection networks by Tripathy et al. [19]. Tsolakis et al. [20] presented a fast fuzzy vector quantization technique for compression of gray scale images. Based on a crisp relation, an input block is assigned to more than one code-word following a fuzzy vector quantization method [21]. LBG Algorithm alone cannot guarantee optimum results since the initialization of the codebook has a significantly impact on the quality of results. Thus, an attempt has been made in this respect where an Improved PSO-LBG (IPSO-LBG) applied to design codebook for Vector Quantization showed fascinating results when compared to other algorithms like QPSO-LBG and PSO-LBG [22]. Contextual region is encoded giving high priority with high resolution Contextual vector quantization (CVQ) algorithm [23]. Huanga et al. devised a dynamic learning vector– scalar quantization for compressing ECG images [24]. Bat Algorithm (BA) is yet another advanced nature inspired algorithm which is based on the echolocation behavior of bats with variation in the pulse rates of loudness and emission. BA in association with LBG is applied as BA-LBG for fast vector quantization in image compression [25]. Results obtained from BA-LBG when compared with QPSO-LBG, HBMO-LBG, PSO-LBG and FA-LBG showed that BA-LBG outperforms the other mentioned algorithms.

Optimization problems are applied in various domains of Engineering Design, Structural Optimization, Economics and Scheduling Assignments. These problems have some mathematical models and Objective Functions. Two varieties of such problems exist: Unconstrained (without constraints) and Constrained (with Constraints) involving both continuous as well as discrete variables. They are usually non-linear in nature. The task of finding the optimal solutions is difficult with several restraints being active at the global optima. Traditional methods for solving these problems are Gradient Descent, Dynamic Programming and Newton Methods. But they are computationally inefficient. This mark the advent of the meta-heuristic algorithms which provide feasible solutions in a reasonable amount of time. There list of meta-heuristics include Genetic Algorithm (GA) [26], Particle Swarm Optimization (PSO) [27], Gravitational Search Algorithm (GSA) [28], Ant Colony Optimization (ACO) [29, 30], Stimulated Annealing (SA) [31, 32], Plant Propagation Algorithm (PPA) [33, 34] and so on [35, 36].

Differential Evolution (DE) is one such metaheuristic algorithm used for fast convergence and less computation time. It optimizes a problem by trying to improve a candidate solution through generations based on a fitness function and using a crossover strategy which is different from that of the Genetic Algorithm. DE is modified to Improved DE (IDE) and used to generate high quality codebook with LBG for vector quantization for image compression. A comparison with other algorithms like IPSO, BA and FA clearly reveals that IDE outperforms all of the other mentioned algorithm in terms of efficiency as judged from the PSNR values of the images.

The paper is organized as follows: In Section 2, codebook design using generalized LBG Vector Quantization Algorithm is discussed. In Section 3, the proposed method of IDE-LBG Algorithm is presented. Section 4 includes the experimental results, discussions and comparisons. Eventually, Section 5 contains the concluding part.

## 2. Codebook Design for Vector Quantization

The Vector Quantization (VQ) is a block (vector) coding technique which is to be optimized for image compression. Designing a proper codebook lessens the distortion between reconstructed image and original image with a modicum of computational time. Let us consider an Image of size $N$ x $N$ which is to be vector quantized. It is then subdivided into $N_b$ blocks with size $n \times n$ pixels, where, $N_b = (\frac{N}{n} * \frac{N}{n})$. These subdivided image blocks are known as training vectors each of which are of size $n \times n$ pixels are represented with $X_i$ (where $i$ = 1, 2, 3,. . ., $N_b$). In the process of Codebook design, a Codebook has a set of code words, where the $i^{th}$ Code-word is represented as $C_i$ (where $i$ = 1, 2, ..., $N_c$) where $N_c$ is the total number of Code-words in Codebook. Every subdivided image vector is approximately represented by the index of a Code-word using a closest match technique. The approach is based on the minimum Euclidean distance between the vector and corresponding code-words. In the decoding phase, the same codebook is used to translate the index back to its corresponding Code-word for image reconstruction. Then the distortion between the actual and the reconstructed image is calculated. The distortion $D$ between training vectors and the codebook is given as:

$$D = \frac{1}{N_c}\sum_{j=1}^{N_c}\sum_{i=1}^{N_b} u_{ij} \cdot \|X_i - C_j\|^2 \qquad (1)$$

Subject to the following restraints:

$$D = \sum_{j=1}^{N_c} u_{ij} = 1, \quad \forall\, i \in \{1, 2, \ldots, N_b\} \qquad (2)$$

$$u_{ij} = \begin{cases} 1, & \text{iff } X_i \in j^{th} \text{ cluster} \\ 0, & \text{otherwise} \end{cases} \qquad (3)$$

The other two necessary conditions which are to be followed for an optimal vector quantizer are:

- The partition $R_j\ \forall\, j = 1,2,3,\ldots,N_c$ must meet the following criterion:
$$R_j \supset \{x \in X : d(c, C_j) < d(x, C_k),\ \forall\, k \neq j\} \qquad (4)$$
- The Code-word $C_j$ should be defined as the centroid of $R_j$ such that,
$$C_j = \frac{1}{N_j}\sum_{i=1}^{N_j} x_i, \quad \forall\, x_i \in R_j \qquad (5)$$

where $N_j$ is the net number of vectors which belong to $R_j$. The procedure for the most commonly used method for codebook design in Vector Quantization, the Lloyd-Buzo-Gray (LBG) Algorithm is provided as follows:

Steps:

1. Initialization: Iteration counter is set as $m = 1$ and the initial distortion $D_1 = 0$. Initial codebook $C_1$ is of size $N$.
2. One block (vector) from the set of training vectors is selected and compared with all code-words from codebook, $C_{m-1} = \{Y_i\}$ to find the closest code-word using nearest neighbor condition and then the code-word set is added.
3. Once all the blocks are grouped into sets of code-words using the above mentioned step, the centroids of all the code-word sets are evaluated to give an improved codebook $C_m$.
4. Next, the total distortion of all code-word sets from the respective centroid points are calculated. The average distortion is given as $D_m$.
5. If $|(D_{m-1} - D_m)| \leq \varepsilon$, the algorithm terminates and the current codebook is recorded as the final answer. Otherwise, $m = m + 1$, the current codebook is used as the initial codebook for step 1 and the entire process is then repeated.

LBG Algorithm ensures the reduction in distortion with increase in iteration number. But there is no assurance regarding the resulting codebook, whether it will become an optimum one or not. It is seen that the initial condition of the codebook has a significant effect on the results. So the initial codebook selection is of prime importance in the LBG Algorithm. Thus we propose to combine an efficient modified heuristic Improved Differential Evolution (IDE) with LBG Algorithm, thereby giving a state-of-the-art IDE-LBG Algorithm and the problem initial codebook selection is tackled with this attempt.

### 3. Proposed IDE-LBG Vector Quantization A:gorithm

Differential Evolution (DE) is a powerful population-based efficient global optimization technique which was posited by Storn et al. [37]. It has been successfully applied to diverse fields including pattern recognition, communication and so on. In our algorithm we have used the DE/current to best/1 scheme. We have modified DE to

IDE with certain modifications in the mutation strategy and in the boundary control strategy. In the mutation strategy we have considered a different scaling factor. The boundary control strategy is improved such that when the limits are crossed then based on a probability there can be either of the two possibilities: (i) the solutions are adjusted within the boundary limits by making the values equal to the values of the limits, or, (ii) by randomly generating a fresh candidate within the bounds and replacing the one which violated the limits.

The training image vectors or blocks are demarcated into different groups. One block from each group is selected at random as the initial candidates. A candidate is constructed by $N_c \times 16$ pixels, where $N_c$ is the size of a codebook. In the process, all the candidates of the entire population are moved nonlinearly to cover all searching space, a process known as exploration. In each generation, $G$, IDE uses the mutation and crossover operations to engender a trial vector $U_{i,}^G$ for each individual vector $X_i^G$, known as target vector in the present population. The mutation and crossover operations along with the fitness function are described below:

   a. **Mutation:**
   There exists an associated mutant vector, $V_{i,}^G$ for each target vector $X_i^G$ at generation $G$ in the current population. The DE/current to best/1 scheme dictates the mutation operations as follows:
   $$V_i^G = X_i^G + F \cdot (X_{best}^G - X_i^G) + F \cdot (X_{r_1}^G - X_{r_2}^G) \tag{6}$$
   where $r_1$ and $r_2$ are random integers which are essentially mutually different and lies in the range [1, $NP$]. $F$ is called the Weighting Factor, $F = 3.randn$, where $randn \sim N(0,1)$, $N$ being the standard normal distribution. $X_{best}^G$ is by far the fittest individual in the population at generation $G$.

   b. **Crossover Operation:**
   Post-mutation period, a binomial crossover strategy is adopted to generate a trial vector $U_{i,}^G = \{u_{1i}^G, u_{2i}^G, \ldots, u_{di}^G,\}$, where $d$ is the dimension. $CR$ is the crossover operator, a value between 0 and 1. We have considered $CR = 0.9$.
   $$U_{ji}^G = \begin{cases} v_{ji}^G, & if\ (rand_j(0,1) \leq CR, or,\ j = j_{rand} \\ x_{ji}^G, & otherwise \end{cases} \tag{7}$$

   c. **Fitness function:**
   Size of the testing images is $N \times N$, $I$ is the recompressed image pixels, and $\bar{I}$ is the compressed image pixels. The fitness function is given in the following set of equations (8-10). First Mean Square Error (MSE) is calculated, then Peak Signal to Noise Ratio (PSNR) is evaluated which is nothing but the fitness function of the problem.
   $$MSE = \frac{\sum_{i=1}^{N}\sum_{j=1}^{N}(I_{i,j} - \bar{I}_{i,j})^2}{N \times N} \tag{8}$$

   $$PSNR = 10 \times \log(\frac{255^2}{MSE}) \tag{9}$$

   $$f_{fitness} = \max(PSNR) \tag{10}$$

The detailed steps for IDE-LBG Algorithm for finding optimal Codebook are described as follows:

1. Parameters and Functions: Fitness function = $f_{fitness}$, iteration parameter = $iter = 1$, number of generations = $N_{Gen}$, codebook size = $N_c$, and population size = $NP$.
2. Testing images divided into $N_b$ non-overlapping image blocks or input vectors ($b_k$ ; $k = 1, 2, \ldots, N_b$). They are put into a training set *tSet*, which of each block is $n \times n$ pixels, the total pixels of each image block is evaluated and the blocks are sorted in the increasing order of pixel number. Here we have considered $n = 4$.
3. $N_b$ blocks are separated into $N_c$ groups according to the index number of blocks, each group having $\frac{N_b}{N_c}$ number of blocks. Blocks from groups are selected at random to form the Code-words to form the initial population of size $NP$.

4. At current iteration *G*, the individual candidates of the population are updated using the mutation and crossover strategies using equations (6)-(7). Boundary control is applied so that the solutions remain within the limits of the search space.
5. The fitness of the off-springs are calculated and compared. If any individual is better than $X_{best}^{G}$, $X_{best}^{G+1}$ is updated with value equal to that of the fittest individual, otherwise $X_{best}^{G+1} = X_{best}^{G}$.
6. If current value of iteration parameter *iter* reaches $N_{Gen}$, then $X_{best}^{Gen}$ is considered as the optimal solution obtained to be used as an initial codebook for the LBG algorithm, otherwise *iter* = *iter* + 1 and the process is repeated again from step 4.
7. Global best candidate solution $X_{best}^{Gen}$ obtained from IDE is used as an initial codebook for LBG Algorithm where each Code-word essentially represents the centroid point of current Code-word set.
8. The LBG Algorithm is then applied whose steps are provided in detail in the previous section (Section 2).

## 4. Experimental Results

In this section, we performed computational experiments and compared the results to show the efficacy of our newly propounded IDE-LBG approach for VQ for Image Compression. All the simulations are carried out in MATLAB 2013a in a workstation with Intel Core i3 2.9 GHz processor. Five tested 512 × 512 grayscale images, namely, ''LENA'', ''PEPPER'', ''BABOON'', ''GOLDHILL'' and ''LAKE'', were selected to be used as training images to train different size of Codebooks and as the basis of comparison of proffered IDE-LBG with other algorithms. The selected images are compressed with IDE-LBG, IPSO-LBG, BA-LBG and FA-LBG and the results are compared. The testing image which is to be compressed is subdivided into non-overlapping image blocks of size 4 × 4 pixels. Each subdivided image block is a training vector of (4 x 4) 16 dimensions. The training vectors to be encoded are ($\frac{512}{2} \times \frac{512}{2}$) 16384 in number. The subdivided image blocks are sorted by the total pixels of each image block and are grouped by the codebook size. Six different codebook sizes are used for comparison with values 8, 16, 32, 64, 128, 256. The values considered are: $\varepsilon$ = 0.001, *NP* = 20, are $N_{Gen}$ = 10, for the training process. The members of the initial population of IDE is chosen at random from each group. Eventually, the best solution (Codebook) obtained by the IDE is used as the initial Codebook for the LBG algorithm-the problem of LBG with random initialization as mentioned before is overcome. We have considered bitrate per pixel (bpp) to evaluate the data size of the compressed image for different codebook sizes and the Peak Signal to Noise Ratio (PSNR) values are calculated for individual codebook sizes. Then the plots of PSNR values versus corresponding bpp values give a clear picture of the quality of reconstructed images with different algorithms considered, thus giving an idea about the effectiveness of the proposed IDE-LBG Algorithm.

$$bpp = \frac{\log_2 N_c}{k} \quad (11)$$

where $N_c$ is the size of the Codebook and *k* is the size of the non-overlapping image block.

The total number of runs considered is 10. Table 1 shows the average PSNR values in ten runs (averaged) obtained by the IDE-LBG, IPSO-LBG, BA-LBG, and FA-LBG algorithms for different codebook sizes ($N_c$ = 8, 16, 32, 64, 128 and 256). Horng et al. [18] simulated five different algorithms in C++6.0, with population size = 100 and number of iterations = 100. PSNR values obtained by different algorithms are also reported in the same paper, but since the platform and the value of parameters, namely, the iteration number and the population size is different, there will be some dissimilarity in the final results obtained for PSNR values of FA-LBG in this paper with that obtained in [17]. The results from the Table 1 (plots are in Fig. 3) confirmed that the fitness (PSNR) of the five test images using the IDE-LBG algorithm is higher than the IPSO-LBG, BA-LBG, and FA-LBG algorithm, yet the difference in PSNR values of IDE-LBG and IPSO-LBG is not very distinct for most of Codebook sizes for various images. Five reconstructed images of Codebook size 128 and block size 16 are also displayed in Fig. 2. It is clearly visible that the reconstructed image quality obtained using the IDE-LBG algorithm has the edge over the IPSO-LBG, BA-LBG and FA-LBG algorithms.

**Table 1.** PSNR values of image compression for 5 different 512 x 512 images with 6 different Codebook sizes.

| Nc | Method | Lena | Baboon | Goldhill | Pepper | Barbara |
|---|---|---|---|---|---|---|
| 8 | IDE-LBG | 25.82 | 20.89 | 26.97 | 25.79 | 25.32 |
|   | IPSO-LBG | 25.75 | 20.33 | 26.12 | 25.78 | 25.27 |
|   | BA-LBG | 24.20 | 19.19 | 25.01 | 24.72 | 24.35 |
|   | FA-LBG | 24.01 | 19.17 | 24.38 | 24.64 | 23.58 |
| 16 | IDE-LBG | 27.19 | 21.11 | 27.32 | 26.51 | 26.03 |
|   | IPSO-LBG | 27.03 | 21.02 | 27.11 | 26.40 | 25.96 |
|   | BA-LBG | 25.67 | 20.17 | 26.23 | 25.95 | 24.88 |
|   | FA-LBG | 25.42 | 20.14 | 25.89 | 25.31 | 24.49 |
| 32 | IDE-LBG | 28.50 | 21.28 | 28.83 | 27.61 | 27.58 |
|   | IPSO-LBG | 28.38 | 21.23 | 28.14 | 27.23 | 27.55 |
|   | BA-LBG | 25.89 | 20.24 | 26.39 | 26.58 | 26.38 |
|   | FA-LBG | 25.73 | 20.26 | 26.36 | 26.53 | 26.36 |
| 64 | IDE-LBG | 29.39 | 22.84 | 28.62 | 29.25 | 28.18 |
|   | IPSO-LBG | 29.32 | 22.72 | 28.44 | 28.95 | 28.07 |
|   | BA-LBG | 26.47 | 21.86 | 27.26 | 28.69 | 27.98 |
|   | FA-LBG | 26.22 | 21.74 | 27.24 | 28.21 | 27.85 |
| 128 | IDE-LBG | 30.45 | 24.09 | 29.92 | 30.84 | 29.94 |
|   | IPSO-LBG | 30.45 | 24.02 | 29.80 | 30.61 | 29.92 |
|   | BA-LBG | 28.23 | 22.43 | 28.30 | 29.90 | 29.01 |
|   | FA-LBG | 27.75 | 22.21 | 28.18 | 29.71 | 28.98 |
| 256 | IDE-LBG | 31.47 | 24.75 | 30.63 | 31.66 | 30.48 |
|   | IPSO-LBG | 31.46 | 24.66 | 30.62 | 31.62 | 30.41 |
|   | BA-LBG | 28.94 | 22.92 | 29.19 | 30.54 | 29.34 |
|   | FA-LBG | 28.49 | 22.58 | 28.93 | 30.44 | 29.23 |

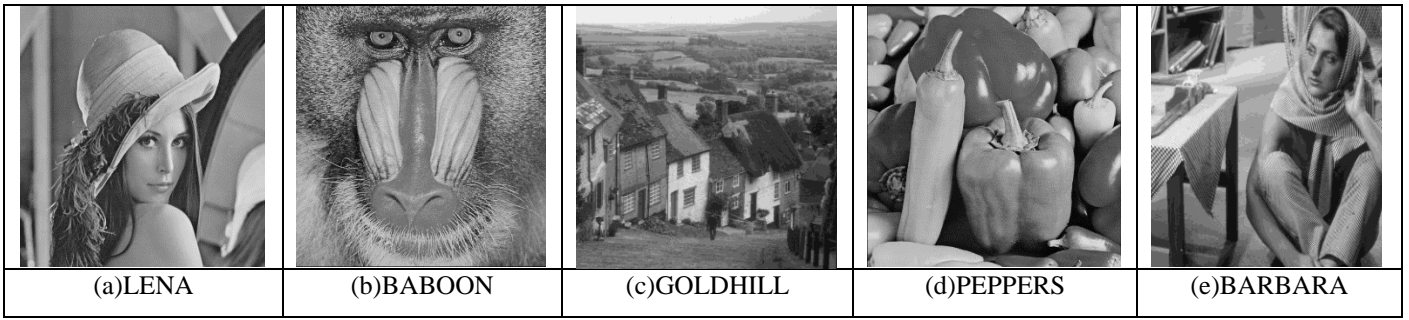

(a)LENA  (b)BABOON  (c)GOLDHILL  (d)PEPPERS  (e)BARBARA

**Fig1.** The five testing Images of size 512 x 512.

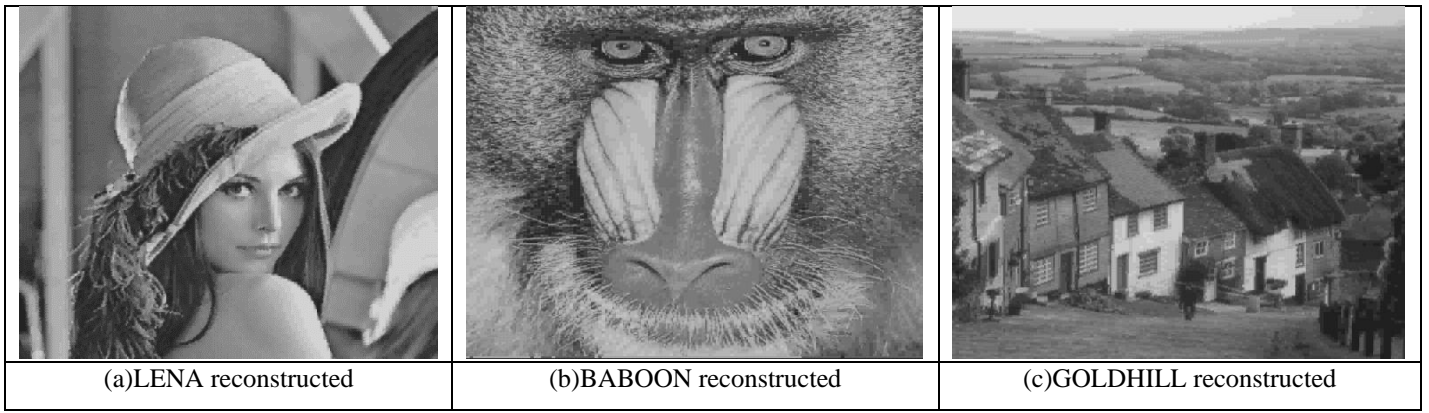

(a)LENA reconstructed  (b)BABOON reconstructed  (c)GOLDHILL reconstructed

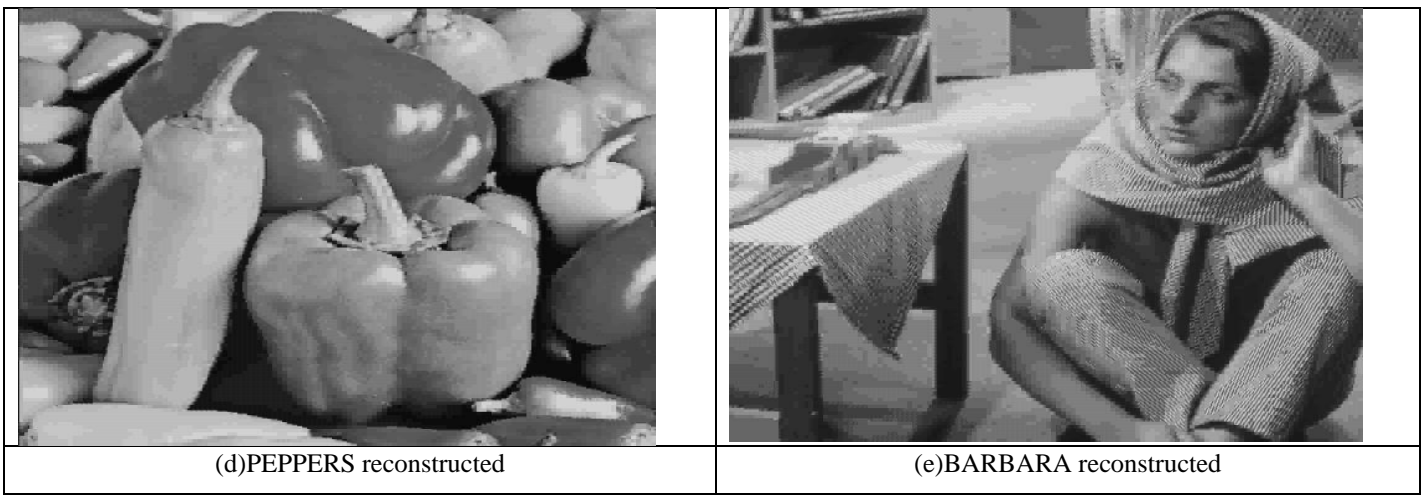

**Fig2.** The five reconstructed Images of Codebook size 128 and Block size 16.

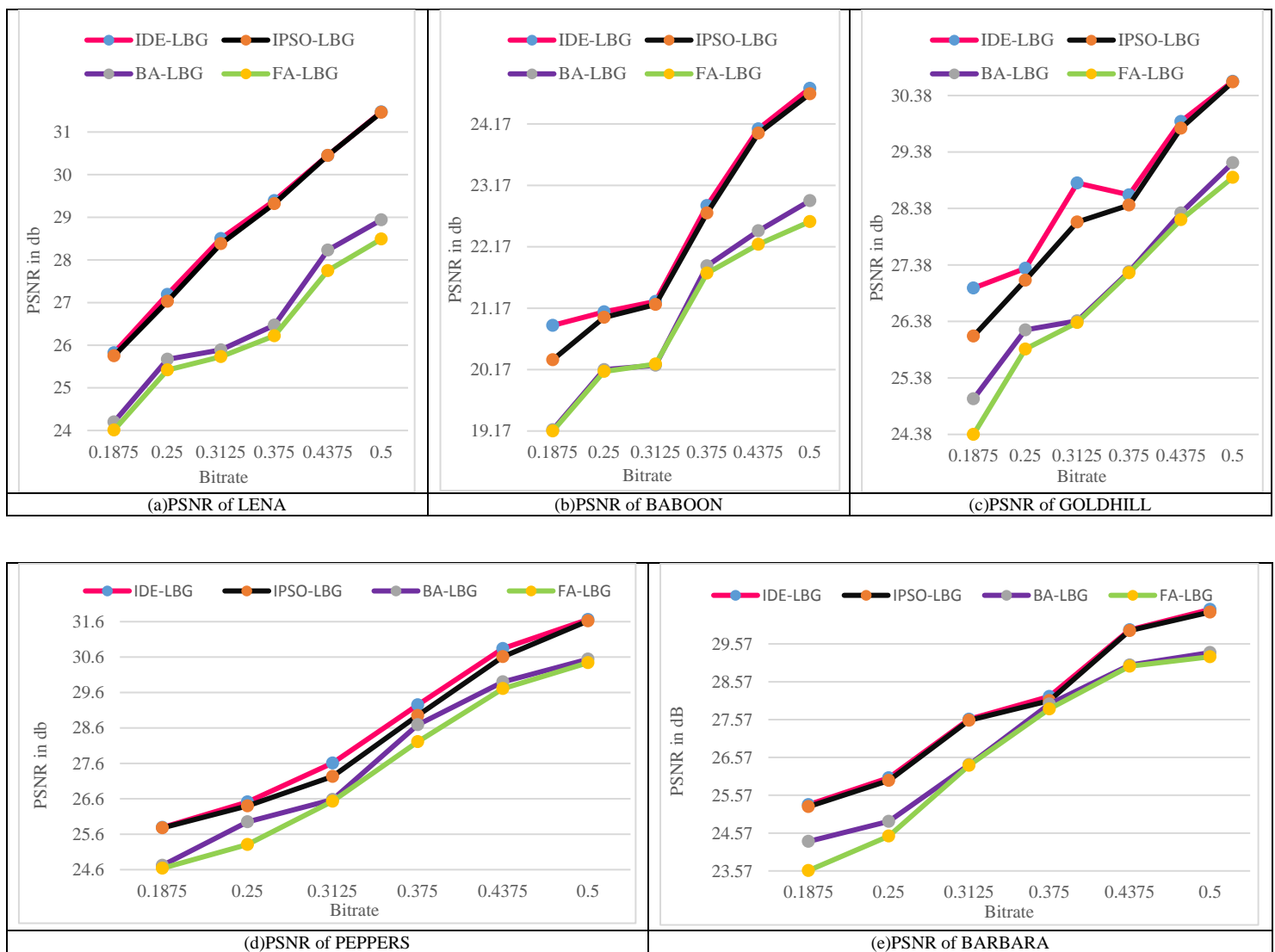

**Fig 3.** Average PSNR of six VQ methods for 5 test images of size 512 x 512

## 5. Conclusion

In this study, an Improved Differential Evolution (IDE) Algorithm based codebook training has been presented for Image Compression. The Peak signal to noise ratio (PSNR) of vector quantization is optimized by using IDE-LBG Algorithm where PSNR is considered as the fitness function for the optimization problem. The algorithm parameters have been tuned suitably for efficient codebook design. The proposed IDE-LBG Algorithm is tested on five standard 512 x 512 grayscale images. Results reveal the potency of the proposed algorithm when compared to IPSO-LBG, BA-LBG and FA-LBG Algorithms. It is observed that using the proposed IDE-LBG Algorithm the PSNR values and the quality of reconstructed image obtained are much better than that obtained from the other algorithms in comparison for six different Codebook sizes.

## References

[1] R.M. Gray, Vector quantization, IEEE Signal Process. Mag. 1 (2) (1984) 4–29.

[2] D. Ailing, C. Guo, An adaptive vector quantization approach for image segmentation based on SOM network, Neurocomputing 149 (2015) 48–58.

[3] H.B. Kekre, Speaker recognition using vector quantization by MFCC and KMCG clustering algorithm, in: IEEE International Conferences on Communication, Information & Computing Technology (ICCICT), IEEE, Mumbai, 2012, pp. 1–5.

[4] C.W. Tsai, C.Y. Lee, M.C. Chiang, C.S. Yang, A fast VQ codebook generation algorithm via pattern reduction, Pattern Recognit. Lett. 30 (2009) 653–660.

[5] S.K. Frank, R.E. Aaron, J. Hwang, R.R. Lawrence, Report: a vector quantization approach to speaker recognition, AT&T Tech. J. 66 (2) (2014) 14–16.

[6] C.H. Chan, M.A. Tahir, J. Kittler, M. Pietikäinen, Multiscale local phase quantization for robust component-based face recognition using kernel fusion of multiple descriptors, IEEE Trans. Pattern Anal. Mach. Intell. 35 (5) (2013) 1164–1177.

[7] G.E. Tsekouras, D. Darzentas, I. Drakoulaki, A.D. Niros, Fast fuzzy vector quantization, in: IEEE International Conference on Fuzzy Systems (FUZZ), IEEE, Barcelona, 2010, pp. 1–8.

[8] Y. Linde, A. Buzo, R.M. Gray, An algorithm for vector quantizer design, IEEE Trans. Commun. 28 (1) (1980) 702–710.

[9] G.E. Tsekouras, D.M. Tsolakis, Fuzzy clustering-based vector quantization for image compression, in: A. Chatterjee, P. Siarry (Eds.), Computational Intelligence in Image Processing, Springer Berlin, Heidelberg, 2012, pp. 93–105.

[10] G. Patane, M. Russo, The enhanced LBG algorithm, Neural Netw. 14 (2002) 1219–1237.

[11] A. Rajpoot, A. Hussain, K. Saleem, Q. Qureshi, A novel image coding algorithm using ant colony system vector quantization, in: International Workshop on Systems, Signals and Image Processing, Poznan, Poland, 2004, pp. 13–15.

[12] C.W. Tsaia, S.P. Tsengb, C.S. Yangc, M.C. Chiangb, PREACO: a fast ant colony optimization for codebook generation, Appl. Soft Comput. 13 (2013) 3008–3020.

[13] M. Kumar, R. Kapoor, T. Goel, Vector quantization based on self-adaptive particle swarm optimization, Int. J. Nonlinear Sci. 9 (3) (2010) 311–319.

[14] H.M. Feng, C.Y. Chen, F. Ye, Evolutionary fuzzy particle swarm optimization vector quantization learning scheme in image compression, Expert Syst. Appl. 32 (2007) 213–222.

[15] Y. Wang, X.Y. Feng, X.Y. Huang, D.B. Pu, W.G. Zhou, Y.C. Liang, et al., A novel quantum swarm evolutionary algorithm and its applications, Neurocomputing 70 (2007) 633–640.

[16] S. Yang, R. Wu, M. Wang, L. Jiao, Evolutionary clustering based vector quantization and SPIHT coding for image compression, Pattern Recognit. Lett. 31 (2010) 1773–1780.

[17] M.H. Horng, Vector quantization using the firefly algorithm for image compression, Expert Syst. Appl. 39 (1) (2012) 1078–1091.

[18] M.H. Horng, T.W. Jiang, Image vector quantization algorithm via honey bee mating optimization, Expert Syst. Appl. 38 (3) (2011) 1382–1392.

[19] P.K. Tripathy, R.K. Dash, C.R. Tripathy, A dynamic programming approach for layout optimization of interconnection networks, Eng. Sci. Technol. 18 (2015) 374–384.

[20] D. Tsolakis, G.E. Tsekouras, A.D. Niros, A. Rigos, On the systematic development of fast fuzzy vector quantization for gray scale image compression, Neural Netw. 36 (2012) 83–96.

[21] G.E. Tsekouras, A fuzzy vector quantization approach to image compression, Appl. Math. Comput. 167 (1) (2005) 539–5605.

[22] Ping-Yi C, Tsai JT, Chou JH, Ho WH, Shi HY, Chen SH. Improved PSO-LBG to design VQ codebook. InSICE Annual Conference (SICE), 2013 Proceedings of 2013 Sep 14 (pp. 876-879). IEEE.

[23] S.M. Hosseini, A. Naghsh-Nilchi, Medical ultrasound image compression using contextual vector quantization, Comput. Biol. Med. 42 (2012) 743–750.

[24] B. Huanga, Y. Wanga, J. Chen, ECG compression using the context modeling arithmetic coding with dynamic learning vector–scalar quantization, Biomed. Signal Process. Control 8 (2013) 59–65.

[25] Chiranjeevi Karri, Umaranjan Jena, Fast vector quantization using a Bat algorithm for image compression, In Engineering Science and Technology, an International Journal, Volume 19, Issue 2, 2016, Pages 769-781, ISSN 2215-0986, https://doi.org/10.1016/j.jestch.2015.11.003.

[26] S.-J. Wu and P.-T. Chow, Genetic algorithms for nonlinear mixed discrete-integer optimization problems via meta-genetic parameter optimization, Engineering Optimization, vol. 24, no. 2, pp. 137–159, 1995.

[27] R. Eberhart and J. Kennedy, A new optimizer using particle swarm theory, in Proceedings of the 6th International Symposium on Micro Machine and Human Science (MHS '95), pp. 39– 43, IEEE, Nagoya, Japan, October 1995.

[28] Rashedi, E., Nezamabadi-pour, H., Saryazdi, S., GSA: A Gravitational Search Algorithm 179(13), 2232–2248 (2009).

[29] M. Dorigo and G. D. Caro, Ant algorithms for discrete optimization, Artificial Life, vol. 5, no. 3, (1999), pp. 137-172.

[30] M. Dorigo and L. M. Gambardella, Ant colony system: a cooperative learning approach to the traveling salesman problem, IEEE Transactions on Evolutionary Computation, vol. 1, no. 1, (1997), pp. 53-66.

[31] C. Zhang and H.-P. Wang, Mixed-discrete nonlinear optimization with simulated annealing, Engineering Optimization, vol. 21, no. 4, pp. 277–291, 1993.

[32] E. H. L. Aarts, J. H. M. Korst, and P. J. M. van Laarhoven, Simulated annealing, in Local Search in Combinatorial Optimization, pp. 91–120, 1997.

[33] Nag S. Adaptive Plant Propagation Algorithm for Solving Economic Load Dispatch Problem. arXiv preprint arXiv:1708.07040. 2017 Aug 4.

[34] Nag S. A Type II Fuzzy Entropy Based Multi-Level Image Thresholding Using Adaptive Plant Propagation Algorithm. arXiv preprint arXiv:1708.09461. 2017 Aug 23.

[35] JC Bansal, H Sharma and SS Jadon, Artificial bee colony algorithm: a survey. Int. J. of Advanced Intelligence Paradigms 5.1 (2013): 123-159.

[36] X.S. Yang, Firefly algorithms for multimodal optimization, LNCS, vol. 5792, pp. 169–178. Springer, Heidelberg (2009).

[37] Storn, R., Price, K., Differential evolution-a simple and efficient heuristic for global optimization over continuous spaces. Journal of Global Optimization 11, 341–359 (1997).